\documentclass[sn-mathphys-ay]{sn-jnl}

\usepackage{amsmath,amssymb}

\usepackage{type1cm}

\begin{document}

\title[Article Title]{Rapid Parameter Inference with Uncertainty Quantification for a Radiological Plume Source Identification Problem}

\author*[1]{\fnm{Christopher} \sur{Edwards}}\email{cjedwar3@gmail.com} \email{Orcid: 0009-0000-0639-0992}

\author[1]{\fnm{Ralph C.} \sur{Smith} \email{Orcid: 0000-0001-7434-5712}}

\affil[1]{\orgdiv{Department of Mathematics}, \orgname{North Carolina State University}, \orgaddress{\street{2311 Stinson Dr.}, \city{Raleigh}, \postcode{27695-8205}, \state{NC}, \country{United States}}}

\abstract{In the event of a nuclear accident, or the detonation of a radiological dispersal device, quickly locating the source of the accident or blast is important for emergency response and environmental decontamination. At a specified time after a simulated instantaneous release of an aerosolized radioactive contaminant, measurements are recorded downwind from an array of radiation sensors. Neural networks are employed to infer the source release parameters in an accurate and rapid manner using sensor and mean wind speed data. We consider two neural network constructions that quantify the uncertainty of the predicted values; a categorical classification neural network and a Bayesian neural network. With the categorical classification neural network, we partition the spatial domain and treat each partition as a separate class for which we estimate the probability that it contains the true source location. In a Bayesian neural network, the weights and biases have a distribution rather than a single optimal value. With each evaluation, these distributions are sampled, yielding a different prediction with each evaluation. The trained Bayesian neural network is thus evaluated to construct posterior densities for the release parameters. Results are compared to Markov chain Monte Carlo (MCMC) results found using the Delayed Rejection Adaptive Metropolis Algorithm. The Bayesian neural network approach is generally much cheaper computationally than the MCMC approach as it relies on the computational cost of the neural network evaluation to generate posterior densities as opposed to the MCMC approach which depends on the computational expense of the transport and radiation detection models.}


\maketitle

\section{Introduction}
\label{sec:1}

Radioactive contamination due to a nuclear accident, leaking radioactive waste, or a purposeful act, such as a radiological dispersal device, can have significant consequences. Contamination within a commercial distract of a large city could bring a devastating economic impact and pose a serious health risk to the surrounding population. Therefore, in the event of a nuclear accident, or a detonation of a radiological dispersal device, quickly locating the source of the release, as well as large spots of radioactive contamination, can save significant time and, consequently, money and resources during the environmental decontamination process, as well as prevent further propagation of deposited radioactive material by wind, rain, and even human transport. This work focuses on locating the source location of an instantaneous release of aerosolized radioactive material as well as estimating the amount of radioactive material released.

Bayesian inference of the release location of a Gaussian plume of a contaminant or pollutant has been studied in previous investigations. Bayesian Markov chain Monte Carlo (MCMC) methods were employed in \citep{Borysiewicz} to infer the release location and release rate of a continuous release of a contaminant using concentration measurements recorded by detectors at the same height as the release. MCMC methods were also used in conjunction with mutual information to infer the release location of a continuous release of a chemical pollutant \citep{Kate}. In \citep{Pudykiewicz}, the author solves the adjoint tracer transport equation to estimate the source location of a nuclear release on a global scale using measurements from the global network of radiation monitoring stations. In \citep{Kopka2015,Kopka2016,Kopka2018}, the authors utilized a sequential approximate Bayesian computation (S-ABC) to localize an atmospheric contamination source. The feasibility of utilizing neural networks for the forward transport model in localization algorithms was investigated in \citep{Wawrzynczak2019, Wawrzynczak2020}. In \citep{Edwards}, we developed a neural network method for real-time inference of a single radiological source without quantification of uncertainty.

To address the goal of real-time inference of the release parameters of an instantaneous release, while also quantifying uncertainty, we propose the construction of a Bayesian neural network to infer the release parameters. The objective is to leverage the stochastic nature and speed of the Bayesian neural network to build predictive distributions by sampling the trained Bayesian neural network evaluations. In doing so, the computational cost of our approach depends on the computational expense of the Bayesian neural network evaluation rather than the computational cost of the transport model, as is typical for Bayesian MCMC approaches.

We first construct a regression-based neural network model that predicts the release location and amount of mass released with no quantification of the uncertainty. This neural network provides a baseline for the other neural network models. The second neural network we construct is a categorical classification model, which utilizes the softmax activation function on its outputs to provide a basic estimate of the probability for the source location. This neural network provides a naive measure of the uncertainty. Finally, we construct a Bayesian regression-based neural network to account for the epistemic uncertainty inherent to the weights of the neural network. By accounting for the aleatoric (random) uncertainty inherent to the detector measurements, and propagating this uncertainty through the neural network model, we are able to construct densities for the source location and mass released. We compare the Bayesian neural network densities to those obtained from the Delayed Rejection Adaptive Metropolis Algorithm (DRAM) \citep{HLMS}.

A key difference in this problem, as compared to source localization problems with a continuous release of a contaminant, is the very short time frame for obtaining useful measurements with a fixed detector. In a continuous release, a stationary sensor can provide highly-informative, non-zero measurements over a long period of time. For an instantaneous release, the hazardous material is only released at a single time and hence can quickly be advected past the site of the detector. A challenge inherent to a radioactive release is that there exist multiple dispersions that can yield the same or similar radiation detector measurements due to the incidence of gamma radiation. 

Modeling the atmospheric transport of radioactive material after a radiological attack is generally performed using computational fluid dynamics (CFD) simulations, Gaussian plume models, or Lagrangian puff, or particle models \citep{Hummel}. We note that RIMPUFF \citep{Rimpuff} and HOTSPOT \citep{Hotspot} are commonly used Gaussian puff and plume models used for rapid response to radiological dispersal devices (RDDs) due to their simplicity and fast computation time. Precomputed CFD simulations are also used for emergency response to radioactive dispersal devices in major cities. Reynolds-averaged Navier-Stokes (RANS) and Large eddy simulations (LES) are used for high fidelity computational fluid dynamics simulations of atmospheric dispersion of material in urban domains. To train our neural networks, we require a significant amount of data. Therefore we use a computationally inexpensive analytic model to compute the mean concentration of radioactive material at a given time and location. We compute the transport of an aerosolized Cs-137 contaminant for several release parameters (mass released, source location, wind speeds). We then compute detector responses for an array of stationary radiation detectors from the resulting dispersion of radioactive material using a ray-tracing model. Using the set of detector measurements, along with the wind speeds, we construct the dataset with which we train our neural networks.

We present the transport and radiation sensor models in Section \ref{sec:transport}. The problem domain and setup along with the stationary array of radiation detectors are defined in Section~\ref{sec:domain2}. We provide technical details for the neural network models in Section~\ref{sec:AFOSR_NN} and summarize techniques for Bayesian inference and DRAM in Section~\ref{sec:Bayesian_inf}. The neural network results are discussed and compared with the DRAM results in Section \ref{sec:chap3_dram_results}. Finally, we provide our concluding remarks and provide future research directions in Section \ref{sec:conclusion3}.   

\section{Methodology}
\subsection{Modeling and Sensing of Atmospheric Dispersion} 
\label{sec:transport}
We summarize here a model governing the transport of aerosolized radioactive material after release. To simplify the discussion and numerical implementation, we focus on 2-D transport in the $x-y$ direction. The extension to 3-D is standard. Based on the air dispersion modeling theory in \citep{Visscher}, we employ the advection-diffusion equation
\begin{equation}
\label{eq:advec_diff}
\frac{\partial c}{\partial t}=-\frac{\partial (c U)}{\partial x}-\frac{\partial (c V)}{\partial y}+\frac{\partial}{\partial x} \bigg(k_{x}\frac{\partial c}{\partial x}\bigg)+\frac{\partial}{\partial y} \bigg(k_{y}\frac{\partial c}{\partial y}\bigg)+S(x,y,t).
\end{equation}
Here $c(x,y,t)$ denotes the mean concentration of the pollutant, $U$ and $V$ are the mean wind speeds in the $x$ and $y$ directions, $k_x$ and $k_y$ are the eddy diffusivities, and $S$ is a source term. We assume constant mean wind speeds and eddy diffusivities so \eqref{eq:advec_diff} can be formulated as
\begin{equation}
\label{eq:advec_diff2}
\frac{\partial c}{\partial t}=-U\frac{\partial (c)}{\partial x}-V\frac{\partial (c)}{\partial y}+k_x\frac{\partial^2 c}{\partial x^2}+k_y\frac{\partial^2 c}{\partial y^2}+S(x,y,t).
\end{equation}
The analytic solution for the mean concentration following an instantaneous release of amount $S$ at the location $x=x_c$ and $y=y_c$ in an infinite fluid with stationary homogenous turbulence is
 \begin{equation}
\label{eq:analytic}
c(x,y,t)=\frac{S}{4\pi t(k_x k_y)^{1/2}} \exp \big(-\frac{(x-x_c-U t)^2}{4k_x t}-\frac{(y-y_c-V t)^2}{4k_y t}\big);
\end{equation}
e.g., see \citep{Visscher}. We focus specifically on the setting where $c(x,y,t)$ is the concentration of Cs-137 in grams at $r=(x,y)$ at time $t$. We take the bounds of the domain to be $Ly_1$, $Ly_2$, $Lx_1$, $Lx_2$. To compute radiation detector measurements for a plume or given dispersion of radioactive material, we employ the relation

\begin{equation}
\begin{split}
\label{eq:sensor_int}
\Gamma(r,t, A,\epsilon, \Delta t)=&\text{Pois}\bigg( \int_{Ly_1}^{Ly_2}  \int_{Lx_1}^{Lx_2}(c \cdot SA) \Delta t \epsilon \frac{A}{4\pi ||r-r_S||_2^2} \\
&\cdot \exp\Big(-\mu_a ||r-r_S||_2\Big) dx dy +B\Delta t \bigg)
\end{split}
\end{equation}
given in \citep{RISO}. This expression models the counts detected for a NaI scintillator detector from the dispersion of CS-137 within the domain. Here $SA=3.214 \cdot 10^{12}  \hspace{1mm} \frac{\text{Bq}}{\text{g}}$ is the specific activity of Cs-137, $\Delta t$ is the dwell time of the detector, $B$ is the background radiation, $\epsilon$ is the intrinsic efficiency of the detector, $A$ is the face area of the detector, and $r$ is the location of the detector. We also assume that there are no obstructions between the detector and plume, so that the only attenuation is due to the linear attenuation factor $\mu_a$ of air.

We discretize the domain into $N_x \times N_y$ cells so that the detector response is given by 
\begin{equation}
\label{eq:sensor_approx}
\Gamma=\text{Pois}\bigg(\sum_{i=1}^{N_x \cdot N_y} (c_i \cdot SA) \Delta t \epsilon \frac{A}{4\pi ||r-r_i||_2^2} \exp\Big(-\mu_a ||r-r_i||_2\Big) \Delta x \Delta y+B\Delta t \bigg).
\end{equation}
Here $c_i$ denotes the concentration in cell $i$ and $r_i=(x_i,y_i)$ denotes the location of the center of the cell.

We assume that the detectors are NaI scintillator detectors measuring 662 keV gamma rays from Cs-137. The detectors each have face area $A=0.0058\hspace{1mm} \text{m}^2$ (3 inch $\times$ 3 inch), a dwell time of $\Delta t=0.1$ seconds and an intrinsic efficiency of $\epsilon=0.62$, which is the approximate intrinsic efficiency of a 3 inch $\times$ 3 inch NaI detector for 662 keV gamma rays. We assume a uniform background count rate of $B_i=300$ counts per second.

\subsection{Release and Measurement Domain}
\label{sec:domain2}

\begin{figure}[!ht]
\centering
\includegraphics[width=0.6\textwidth]{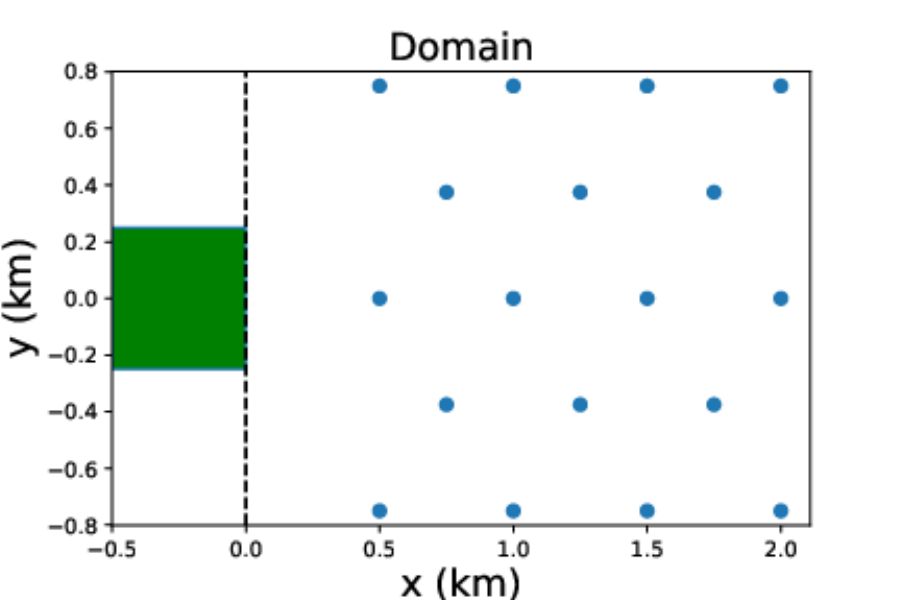}
 \caption{Problem domain where the shaded area denotes the potential release region and dots denote the 18 radiation sensors located downwind from the release.}
 \label{fig:geom_AFOSR}
  \end{figure}

We assume that aerosolized Cs-137 is released from a location inside a $500 \hspace{1mm} \text{m} \times 500 \hspace{1mm} \text{m}$ region, with the amount of radioactive material released ranging from 1 to 5 grams. We further assume that there are 18 NaI scintillator detectors systematically placed within a $2000 \hspace{1mm} \text{m} \times 1500 \hspace{1mm} \text{m}$ region downwind of the release region. We assume that the time of release is known and the detectors record measurements at $t=500$ s after release. We also assume that the mean wind speed for the 500 s after release is known. The release and sensor domain is depicted in Figure \ref{fig:geom_AFOSR}.

We make the following assumptions regarding the problem formulation. We first assume a 2-D rather than 3-D domain to simplify the computation of \eqref{eq:sensor_approx} and avoid the large number of aerial detectors that would be necessary to cover the region for a 3-D problem. We also assume that the time of release is exactly known, which is not true in general. Uncertainty in the release time increases the error of predictions and the width of posterior densities inferred by Bayesian techniques. Finally, we assume that the mean wind speeds are known over a 500 second period and are uniform across the region.

\subsection{Neural Networks}
\label{sec:AFOSR_NN}
For an $n$ layer neural network with multiple outputs, the value of the $k$th output is given by

\begin{equation}
\label{eq:hiddenlayer3}
y_k=\sum_{j=1}^{N_n} W_{jk}^nq_j+b_k^n, \hspace{4mm} k=1, \dots, M.
\end{equation}
Here, $b_k^n$ is the bias associated with the $k$th element of the output, $q_j$ is the output from the $j$th node of the $n$th layer, and $M$ is the dimension of the output. Additionally, $N_n$ is the number of nodes in the $n$th hidden layer and $W_{jk}^n$ denotes the $jk$ entry in the $N_n \times M$ matrix $W^n$ of weights associated with the output. 

For the inverse problem of inferring the release parameters of a release of aerosolized radioactive material, given a set of radiation detector measurements and mean wind speeds, we utilize a neural network model to infer the release parameters. We optimize the weights and biases of the neural network model for the recovery of the release parameters by minimizing the mean absolute error of the predicted release parameters over a dataset $D={x_k, y_k, k=1, \dots, N}$, of detector observations/mean wind speeds and corresponding release parameters that produced the detector observations.\\

\subsubsection{Regression-Based Neural Network}
\begin{figure}[!ht]
\centering
\includegraphics[width=0.7\textwidth]{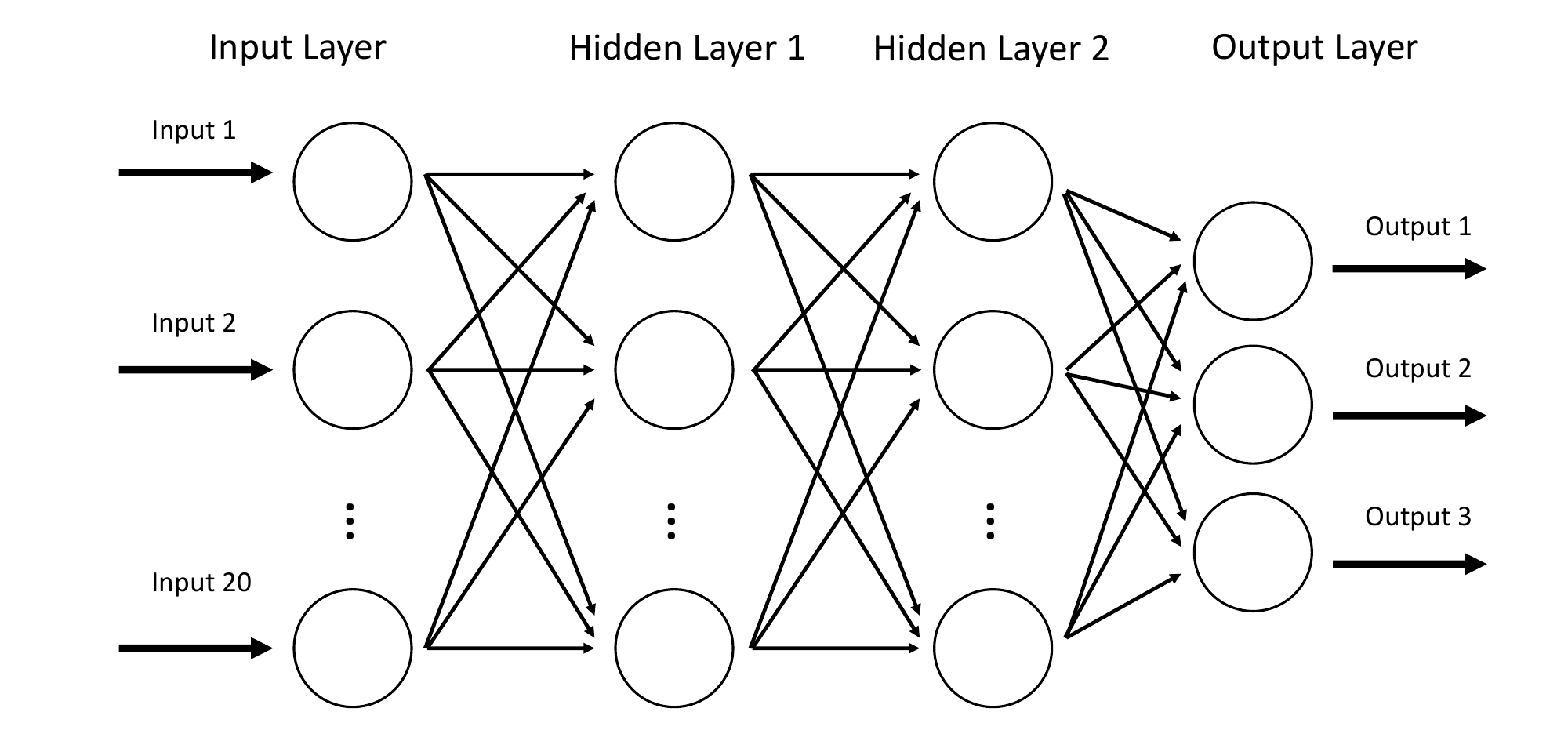}
 \caption{Architecture of the regression-based neural network.}
 \label{fig:NN_diagram}
  \end{figure}

We summarize first the architecture of the regression-based neural network used to infer the release parameters of the source. We employ a 3 layer feedforward neural network. This neural network will serve as a baseline for the development of neural network models that quantify uncertainty. The first two layers each employ 150 nodes and 200 nodes, respectively, and utilize the swish activation function
\begin{equation}
\label{eq:swish2}
\Psi(x)=\frac{x}{1+\exp(-x)}.
\end{equation} 
The output layer consists of 3 nodes and employs a linear activation function. The architecture of this neural network is depicted in Figure \ref{fig:NN_diagram}. \\

\subsubsection{Categorical Classification-Based Neural Network}
In addition to a regression-based neural network, we employ a categorical classification-based neural network with a softmax activation function on the output layers to qualify uncertainty. The motivation for this classification is to separate the parameter domain for the release location into bins and predict which bins are most likely to contain the release location of the source. If we take the bin width to be sufficiently small, then the resulting outputs, which can be interpreted as an approximation of the probability that the true release location falls within a specified bin, can approximate the parameter density. 

We separate the $500$ m $\times 500$ m domain into 100 bins for the $x$ prediction and 100 bins for the $y$ prediction. The architecture of this neural network differs in the output layer. The categorical classification neural network predicts only the release location and the output layer contains 100 nodes for each of the outputs we wish to predict, with 1 node corresponding to each bin. The output layer employs the softmax activation function
\begin{equation}
\label{eq:softmax}
\Psi(x)_i= \frac{\exp(x_i)}{\sum_{j=1}^K\exp(x_j)}.
\end{equation} 
Here $K$ is the number of classes, or bins, and division by $\sum_{j=1}^K\exp(x_j)$ guarantees that the output vector sums to 1. The weights and biases of the categorical classification neural network are trained by minimizing the categorical cross entropy loss function.

\subsubsection{Bayesian Neural Network with Variational Inference}
\label{sec:Bayesian_NN}
To quantify the epistemic uncertainty inherent to the regression-based neural network, we implement a Bayesian neural network \citep{Cinelli}. In a Bayesian neural network, one assumes that the neural network weights $w$ and biases have distributions rather than a single set of optimal values. This posterior distribution for the weights, $p(w | D)$, is inferred from the data $D={x_k, y_k, k=1, \dots, N}$. From Bayes' rule, one notes that
\begin{equation}
\label{eq:Bayes_rule}
p(w | D)=\frac{p(D | w) p(w)}{p(D)}.
\end{equation} 
Here $p(D | w)$ is the likelihood which qualifies how likely the data is given the neural network weights and biases, $p(w)$ is the prior, and $p(D)=\int p(D | w) p(w) dw$ is a normalizing factor. This integral is intractable for neural networks in general, so evaluation or differentiation of the denominator is impossible. Similar difficulties prohibit the direct computation of marginal distributions. Instead, the posterior distribution for the weights is generally computed using a variational approximation $q(w | \Psi)$. In this approach, one assumes that the weights $w_i$ are independent, each represented by a Gaussian distribution $\Psi_i=(\mu_i, \sigma_i^2)$ with separate means $\mu_i$ and variances $\sigma_i^2$. It follows that the approximating variational posterior density has a Gaussian distribution with a diagonal covariance matrix. We additionally assume that the prior for the weights is Gaussian with a diagonal covariance matrix. We also assume zero mean and a standard deviation of 1 in the Gaussian prior of each weight.

The objective is to minimize the Kullback-Leibler divergence between $p(w | D)$ and the variational approximation $q(w | \Psi)$ to optimize the parameters of the variational posterior $\Psi$. Minimizing the Kullback-Leibler divergence between $p(w | D)$ and the variational approximation $q(w | \Psi)$ is equivalent to maximizing the Evidence Lower BOund (ELBO), which is given by

\begin{equation}
\begin{split}
\label{eq:ELBO}
ELBO(\Psi) &= \int q(w | \Psi) \log{p(D | w)}dw - \int q(w |\Psi)\log{\frac{q(w | \Psi)}{p(w)}} dw \\
&=E_{q(w |\Psi)}\big[\log{p(D | w)}\big]-D_{KL}(q(w |\Psi)|| p(w)); 
\end{split}
\end{equation}
see \citep{Cinelli}. The first term in \eqref{eq:ELBO} trains the parameters of the variational approximation to better fit the data, while the second term is independent of the data and guides the variational posterior distribution toward the prior distribution. As a result, when there is little data the second term is more influential and the posterior distribution of the variational parameters stays near the prior distribution.

The parameters of the variational approximation are optimized using a forward pass with stochastic sampling and a backward pass during which backpropagation is performed to update the parameters. In the forward pass, the objective function is evaluated by sampling from the variational approximation. A reparameterization estimator is employed to ensure differentiability for backpropagation \citep{Kingma}. This permits the derivative of the expectation to be expressed as the expectation of the derivative for Gaussian distributions. This reparameterization was extended in \citep{Blundell} to accommodate non-Gaussian distributions. Following the forward pass, the parameters of the variational approximation $\Psi$ are updated via backpropagation.

After optimizing the parameters of the variational approximation, the trained Bayesian neural network can quantify the epistemic uncertainty of the predicted release parameters by leveraging the stochastic nature of the Bayesian neural network. The aleatoric uncertainties of the radiation detector inputs are then propagated through the Bayesian neural network to construct release parameter densities via sampling the Bayesian neural network evaluations. This procedure produces parameter densities consistent with those obtained using MCMC methods; however, the computational cost of this approach is generally much cheaper since it depends on the computational cost of the neural network evaluation rather than the computational expense of the transport and radiation detection models.

\subsubsection{Data Generation and Neural Network Training}
\label{sec:NN_training}
We generate 400{,}000 release locations and masses for the amount of radioactive material released, specified in grams. We then sampled 400{,}000 corresponding mean wind speeds, $u$ and $v$, respectively, which are the mean wind speed in the $x$ and $y$ directions. Each of the datasets were sampled from the following uniform distributions: $x_c \sim U(-500 \hspace{1mm} \text{m},0 \hspace{1mm} \text{m})$,  $y_c \sim U(-250 \hspace{1mm} \text{m},250 \hspace{1mm} \text{m})$, $m_c \sim U(1 \hspace{1mm} \text{g},5 \hspace{1mm} \text{g})$, $u \sim U(2 \hspace{1mm} \text{m/s},4 \hspace{1mm} \text{m/s})$, $v \sim U(-1 \hspace{1mm} \text{m/s},1 \hspace{1mm} \text{m/s})$. We compute the mean concentration of radioactive material 500 s after release using \eqref{eq:analytic} and then compute the resulting radiation detector measurements using \eqref{eq:sensor_approx}. We employ a 50-25-25 split of the data for training, test, and validation datasets.

The inputs for all neural networks were taken to be the mean wind speeds over the 500 s after release and the log of the detector measurements. The logarithm is applied prior to training as the detector measurements may vary by orders of magnitude. The outputs for the regression-based neural network are the predicted source release location ($x$, $y$) and the predicted mass of Cs-137 released. The outputs of the classification neural network are two 100 $\times$ 1 vectors containing the estimated probability that the source release location ($x$, $y$) fall within the given bin.

We utilize a Nadam optimizer, which adaptively changes the learning rates for each weight during learning. It is similar to the Adam optimizer but incorporates the Nesterov momentum rather than classical momentum \citep{Nadam, Hal}. We adopt a learning rate of $0.001$ and use the mean-absolute error loss function for the regression-based neural network. Explicitly, the mean-absolute error loss function is taken to be
\begin{equation}
\label{eq:MAE}
L(Y,\tilde{Y})=\frac{1}{N}\sum_{i=1}^N |Y_i-\tilde{Y}_i|,
\end{equation}
where $Y_i$ is the true value and $\tilde{Y}_i$ is the prediction. This neural network is trained on the training data with a batch size of $128$ for a total of 500 epochs.

For the categorical classification-based neural network, we employ many of the same hyper-parameters for training. We keep the same learning rate and batch size, but adjust the number of epochs to 1000 based on the observation that longer training resulted in more accurate models. The classification neural network employs the categorical cross entropy loss function

\begin{equation}
\label{eq:CCE}
CCE(Y,\tilde{Y})=-\frac{1}{N} \sum_{i=1}^N \sum_{j=1}^K Y_i^j \log{\tilde{Y_i^j}}.
\end{equation}

\subsection{Bayesian Inference}
\label{sec:Bayesian_inf}
To provide a framework for Bayesian inference, we employ the observation model
\[Y_i=f(x_i,\theta)+\varepsilon_i\]
where $Y_i$ denote random variables representing measurements, $f(x_i,\theta)$ is the parameter dependent model response, $\theta$ are parameters to be inferred, $\varepsilon_i$ are observation errors, and $x_i$ are independent variables which, in this case, are space. We assume that observation errors are independent and identically distributed (iid) with a mean-zero Gaussian distribution and variance $\sigma^2$; i.e. $\varepsilon_i \stackrel{iid}{\sim} N(0,\sigma^2)$. In the Bayesian framework, we consider parameters to be random variables with associated distributions. From a Bayesian perspective, an inverse problem can be stated as follows. Given measurements or data, one wants to find the posterior density that best represents the distribution of parameters.

We construct posterior densities in the following manner. Consider a random parameter vector $\theta$ and known prior density $\pi_0(\theta)$. Let $y$ be a realization of a random observation vector $Y$, which is related to a model via the likelihood $p(y| \theta)$. With the assumptions of Gaussian and iid observation errors the likelihood function $p(y|\theta)$, is given by

\[
    p(y|\theta) = \frac{1}{\left(2\pi\sigma^2\right)^{n/2}} \exp{\Big(\frac{-SS_\theta}{2\sigma^2}\Big)}.
\]
Here
\[
    SS_\theta = \sum_{i=1}^n[y_i-f(x_i,\theta)]^2,
\]
denotes the sum of squares error. The posterior density of $\theta$, given the observations $y$, is given by
\begin{equation}
\label{eq:Bayes}
\pi(\theta | y)=\frac{p(y|\theta) \pi_0(\theta) }{\int_{\mathbb{R}^p} p(y|\theta) \pi_0 (\theta) d\theta}.
\end{equation}
This posterior density $\pi(\theta | y)$ provides a probabilistic estimate of the unknown parameters $\theta$, given the prior density and model fit to the data \citep{kaipio2006,Smith2013Uncertainty}.

Direct evaluation of \eqref{eq:Bayes} is generally not possible due to the computational cost of computing the normalization constant and marginal distribution. This challenge is made more difficult by the fact that the support of the density is often unknown. This motivates the use of Monte Carlo Markov Chain (MCMC) techniques to construct chains whose stationary distribution is the posterior density we seek \citep{Smith2013Uncertainty}. For MCMC techniques, we employ the Delayed Rejection Adaptive Metropolis (DRAM) algorithm detailed in \citep{kaipio2006,Smith2013Uncertainty}.

\section{Inference of Release Parameters}
\label{sec:chap3_dram_results}
\subsection{DRAM Results}
\label{sec:DRAM_results}
We first employ DRAM to compute a posterior distribution for $\theta=[x_c,y_c,m_c]$ and then compare the distribution to the predictions we obtain from the much faster neural network models. We consider a release of 1.83 grams of Cs-137 from $(-389.00 \hspace{1mm} \text{m}, 185.37 \hspace{1mm} \text{m})$ with mean wind speeds $(2.44 \hspace{1mm} \text{m/s},0.74 \hspace{1mm} \text{m/s})$ and eddy diffusivities $(5 \hspace{1mm} \text{m$^2$/s}, 5 \hspace{1mm} \text{m$^2$/s})$. At a time 500 seconds after the release, we record detector measurements from an array of 18 detectors by using \eqref{eq:sensor_approx}.

We form a 1.5 km $\times$ 1.5 km computational grid about the center of the plume at $t=500$~s with 1001 $\times$ 1001 grid points, where the mean concentration is evaluated using \eqref{eq:analytic}. The concentration in each cell is computed using Simpson's rule. After the cell concentrations and centers are computed, the sensor model \eqref{eq:sensor_approx} is used to compute the detector observations from the plume of Cs-137. This computation comprises the bulk ($80\%$) of the computational expense when DRAM calls the model. The transport model and Simpson's rule comprise the remaining $20\%$ of the computational expense.

Using the measurements from this single time snapshot, we employ the DRAM algorithm \citep{HLMS,Smith2013Uncertainty} to estimate parameters $\theta=[x_c, y_c,m_c]$, which is the release location and amount of mass released. A 21 $\times$ 21 $\times$ 20 initial grid search optimization was performed to obtain the parameter values to initialize the chain. We initialize the chain in the DRAM algorithm with the parameters from the grid search optimization that yielded the smallest error. We assume uniform prior distributions over the parameter bounds $x_c \sim U(-500 \hspace{1mm} \text{m},0 \hspace{1mm} \text{m})$,  $y_c \sim U(-250 \hspace{1mm} \text{m},250 \hspace{1mm} \text{m})$, $m_c \sim U(0 \hspace{1mm} \text{g},5 \hspace{1mm} \text{g})$ for the 3 parameters ($x_c$, $y_c$ $m_c$) and we employ the Gaussian likelihood function.

\begin{figure}
    \centering
    \begin{tabular}{c c}
    \includegraphics[width=6cm]{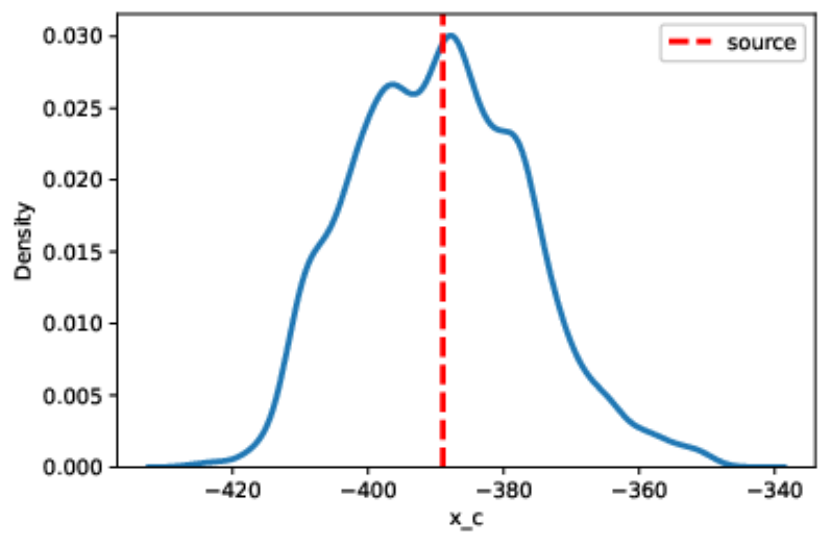}
    \includegraphics[width=6cm]{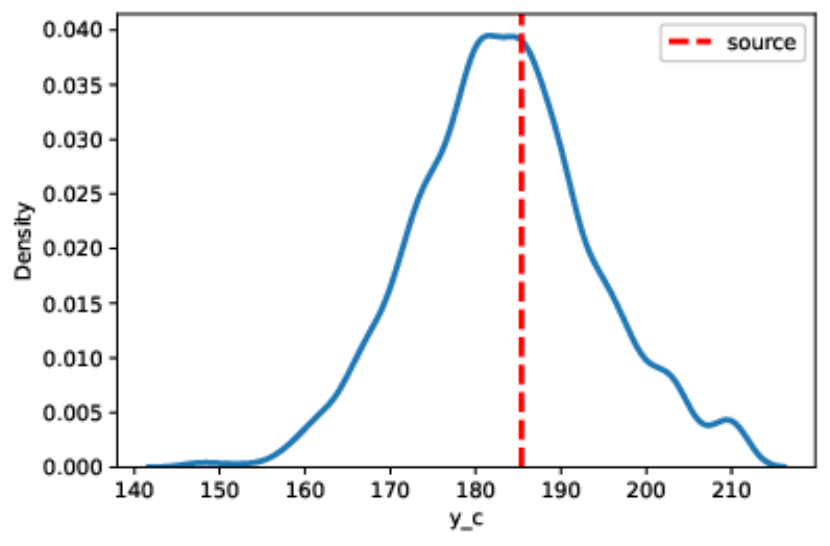}\\
    \includegraphics[width=6cm]{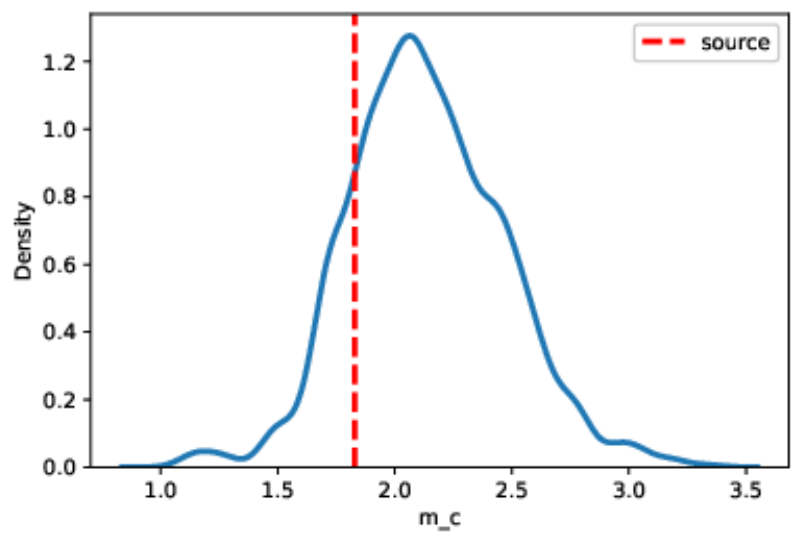}
    \end{tabular}
         \caption{Marginal densities for release parameters $x_c$, $y_c$, and $m_c$.}
    \label{fig:densities}
\end{figure}
\begin{figure}
    \centering
    \begin{tabular}{c c}
    \includegraphics[width=6cm]{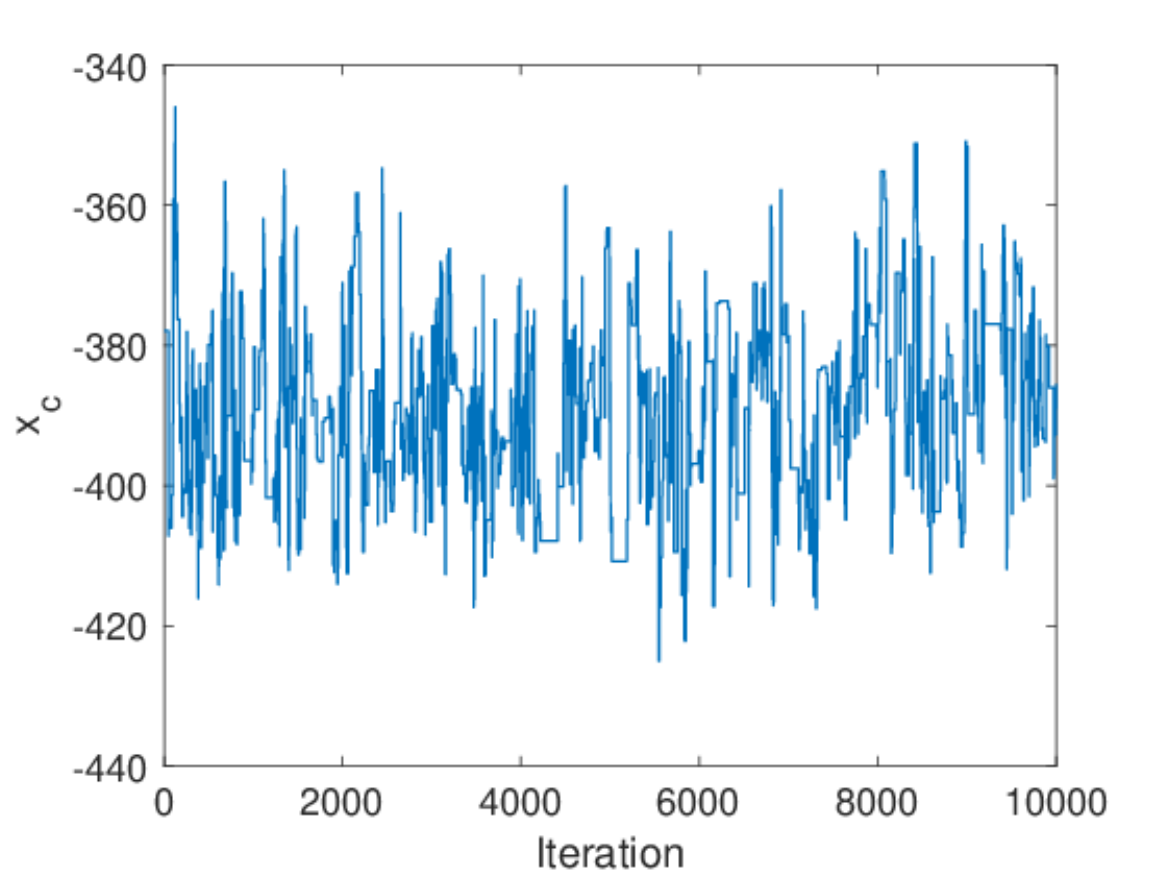}
    \includegraphics[width=6cm]{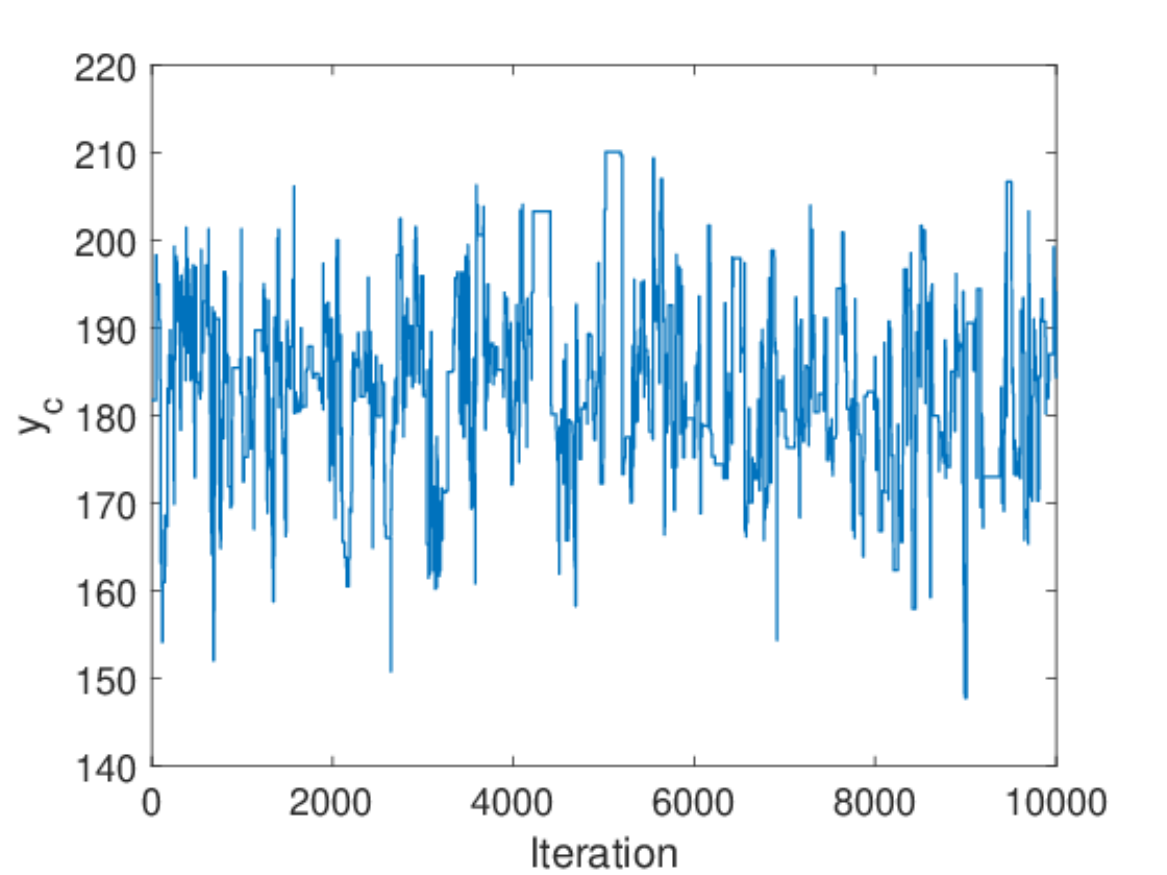}\\
    \includegraphics[width=6cm]{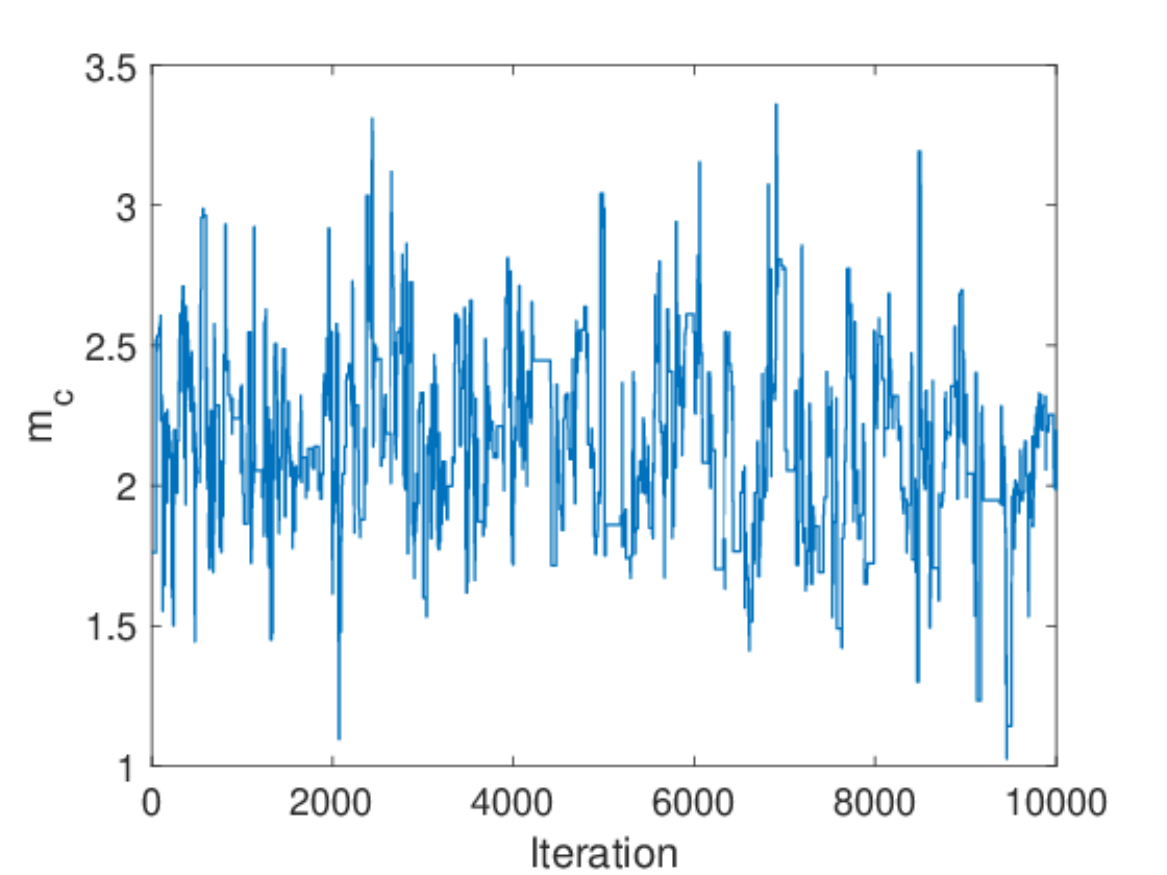}
    \end{tabular}
         \caption{Chains for the release parameters $x_c$, $y_c$, and $m_c$.}
    \label{fig:chains}
  \end{figure}

We ran the DRAM algorithm for 20,000 iterations, taking the first 10,000 iterations as our burn-in, to obtain the parameter densities and corresponding parameter chains. The marginal parameter densities are plotted in Figure~\ref{fig:densities} and the burned-in parameter chains are plotted in Figure~\ref{fig:chains}. 

 \begin{figure}
    \centering
    \includegraphics[width=7cm]{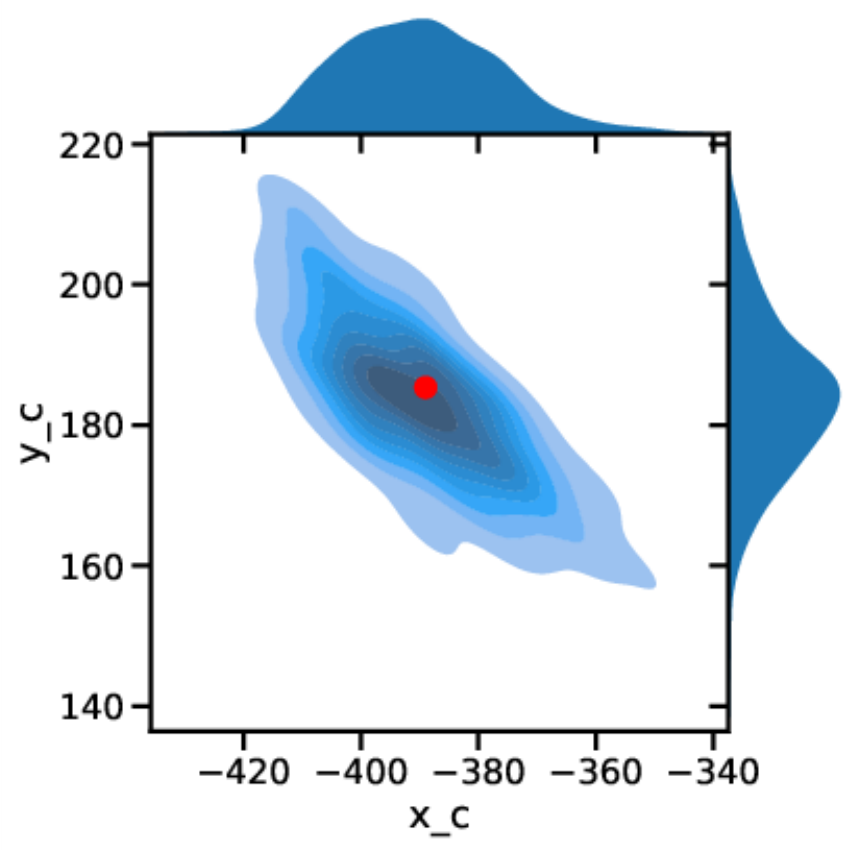}
     \caption{Joint posterior density of release location. The true release location of the instantaneous release is indicated by the red point.}
         \label{fig:posterior}
 \end{figure}

We plot the joint posterior density along with the true source location in Figure \ref{fig:posterior}. We note that the true release parameters are contained within the joint posterior densities that we obtained using DRAM. The mean of the burned in chain for the release parameters is $\hat{x}_c=-389.3 \hspace{1mm} \text{m}$, $\hat{y}_c=183.4 \hspace{1mm} \text{m}$, $\hat{m}_c=2.14$, which is close to the actual release location $x_c=-389.0 \hspace{1mm} \text{m}$ $y_c=185.4 \hspace{1mm} \text{m}$. The mean of the release amount $m_c$ is larger than the true release amount, but it is still contained within the marginal density for $m_c$. For this problem, we found that the initialization of the chain as well as accurate covariance estimates were important for fast convergence and accurate localization of the release parameters. Although accurate, the DRAM algorithm required approximately 1674 seconds to construct posterior densities. 

\subsection{Regression-Based Neural Network Results}
To infer the release parameters $\theta=[x_c,y_c,m_c]$ in real time, we utilize feedforward neural network models. We first consider the regression-based neural network architecture described in Section \ref{sec:AFOSR_NN}. After training on the large amount of generated data, in the manner detailed in Section \ref{sec:NN_training}, we evaluate the accuracy of the neural network and compare the accuracy and speed to the DRAM results. The average error in predicting the release location is $10.44 \hspace{1mm} \text{m}$ and the mean absolute error in predicting the mass of Cs-137 released is $0.2515$ grams. Overall, this average accuracy is very good considering that we are varying the amount of Cs-137 released for each simulation. The evaluation time for a $100{,}000$ large dataset was 0.0625 s for this neural network. The training time for the neural network was 1355 s. 

We next consider the regression neural network's predictions for the specific release parameters examined using DRAM. Given the mean wind speed $(2.44 \hspace{1mm} \text{m/s},0.74 \hspace{1mm} \text{m/s})$ and the 18 detector measurements 500 s after the release of a Cs-137 source, the neural network predicted the source release location $x_c=-384.3 \hspace{1mm} \text{m}$, $y_c=179.6 \hspace{1mm} \text{m}$ and the release mass $m=2.16 \hspace{1mm} \text{g}$, in just $0.00209 \hspace{1mm} \text{s}$. This prediction is less accurate than the means of the chain predicted by DRAM, but it is significantly faster than DRAM which requires 1674 s for this problem. The predicted release location in this case is very close to the actual release location $x_c=-389.0 \hspace{1mm} \text{m}$, $y_c=185.4 \hspace{1mm} \text{m}$ and reasonably close to the actual mass 1.83 g. 
 
We note that although this prediction is accurate, it does not provide qualified uncertainties. If we did not know the true value of the release parameters, we would not know how much to trust the predictions. This illustrates the shortcomings of a regression-based neural network, which outputs a single prediction with no quantification of uncertainty. This motivates the second neural network, a categorical classification neural network which estimates the probability that the release location of the source will fall within the defined bins.

 \subsection{Categorical Classification Neural Network Results}
 To construct a categorical classification network representation, we utilize 100 bins for $x_c$ and 100 bins for $y_c$. Since the domain for the release location is $500 \hspace{1mm} \text{m} \times 500 \hspace{1mm} \text{m}$, the bin-width is $5 \hspace{1mm} \text{m}$ for each parameter. The binning process introduces some error so we expect this neural network to perform slightly worse than the regression-based neural network. The trained categorical classification-based neural network provides a set of estimated probabilities, with each returned value equating to the estimated probability that the true source location occurs within corresponding bin in the $x,y$ domains, respectively. Additionally, we wish to obtain deterministic predictions for evaluating the accuracy of the neural network and for comparison to other methods. To achieve this, we utilize the expectation as a deterministic prediction to evaluate the accuracy of the categorical classification neural network. 
 
 We define the expectation of $\tilde{x}_c^i$ to be
 
\begin{equation}
\label{eq:x_c}
\tilde{x}_c^i= \sum_{j=1}^{100} p_j x_j,
\end{equation} 
where $x_j$ denotes the midpoint of the $j$th bin and $p_j$ denotes the predicted probability that the true $x_c$ occurs within bin $j$. Similarly, we define the expectation of $\tilde{y}_c^i$ as

\begin{equation}
\label{eq:y_c}
\tilde{y}_c^i= \sum_{j=1}^{100} p_j y_j,
\end{equation} 
where $y_j$ denotes the midpoint of the $j$th bin and $p_j$ denotes the predicted probability that the true $y_c$ occurs within bin $j$.

Using these definitions, we compute deterministic values from the categorical classification neural network's predicted probabilities and evaluate the accuracy of the categorical classification neural network using the same criteria used for the regression-based neural network. 

For the test dataset, we record a mean error of $11.42 \hspace{1mm} \text{m}$ when comparing the expectations of the predicted source location from the categorical classification neural network to the true values of the release parameters. As hypothesized, we observe that the mean error is slightly larger than that of the regression-based neural network, but unlike the regression-based neural network, it provides a measure of uncertainty. The training time for the categorical classification neural network was 3547 s and the time required to evaluate the test dataset was 0.141 s.

 \begin{figure}[b]
    \centering
    \begin{tabular}{c c}
    \includegraphics[width=7cm]{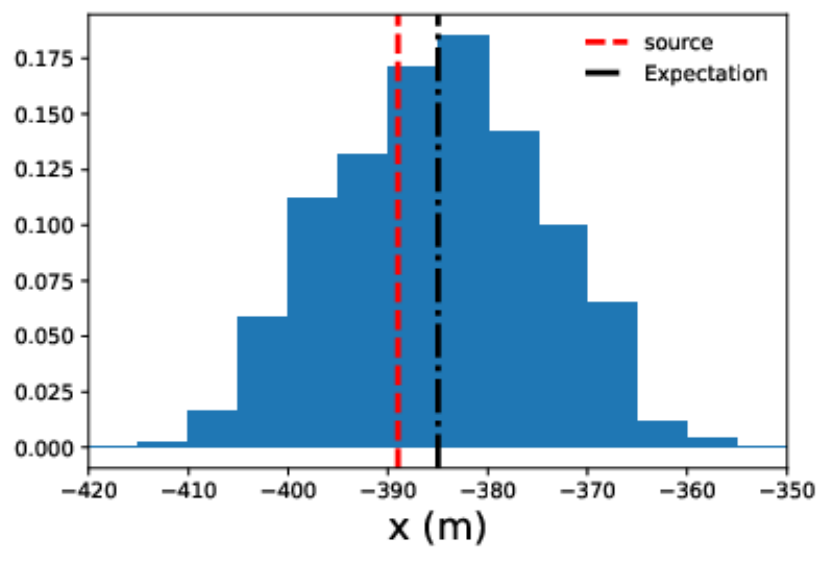}
    \includegraphics[width=7cm]{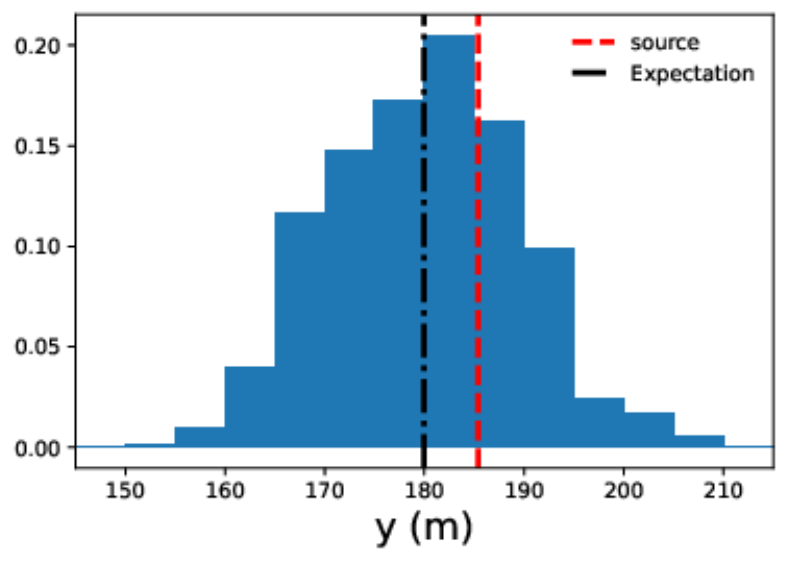}
    \end{tabular}
     \caption{Normalized histogram for the estimated probabilities. The true source release location of the instantaneous release is indicated by the red line and expectation of the neural network prediction is indicated by the black line.}
         \label{fig:prob_x_c}
 \end{figure}

We next apply the categorical classification neural network to the test problem we examined in \ref{sec:DRAM_results}. We now use the categorical classification neural network to predict the source location of the release. The classification neural network computes the estimated probabilities for each bin in $0.00262 \hspace{1mm} \text{s}$, which is slightly slower than the smaller regression neural network ($0.00209 \hspace{1mm} \text{s}$) but still significantly faster than DRAM. The results are plotted in 
Figure~\ref{fig:prob_x_c}.

We note that the classification neural network performs well for this problem since the true source location is contained within or adjacent to the highest probability bins with the computed expectation close to the true values for the source location. We also note that the variance of the estimated probability is very similar to the variance of the DRAM produced densities for $x_c$ and $y_c$. Therefore this approach of uncertainty quantification performs well for this problem.

\subsection{Bayesian Neural Network Results}
To obtain a more robust estimate of the uncertainty than that provided by the categorical classification neural network, we implement a Bayesian neural network in the manner detailed in Section \ref{sec:Bayesian_NN}. The resulting Bayesian neural network's mean error in predicting the source location was $10.22 \hspace{1mm} \text{m}$ over the test dataset for the first inference. Since the weights are sampled from the weight distributions, we obtain different predictions with each evaluation of the Bayesian neural network. Evaluating the mean error for the test dataset 1000 times reveals that the mean accuracy varies little, with the mean error for the predicted source location fluctuating between $10.20 \hspace{1mm} \text{m}$ and $10.28 \hspace{1mm} \text{m}$ at the extremes. Similarly, with the mean absolute error for the mass, we record a mean error of $0.2797 \hspace{1mm} \text{g}$ for the first pass and a range of $0.2787 \hspace{1mm} \text{g}$ to $0.2894 \hspace{1mm} \text{g}$ at the extremes. The Bayesian neural network required 3649 s to train and the Bayesian neural network evaluated the testing dataset in 0.0808 s.

\begin{figure}[!ht]
 \centering
\begin{tabular}{c c}
    \includegraphics[width=6cm]{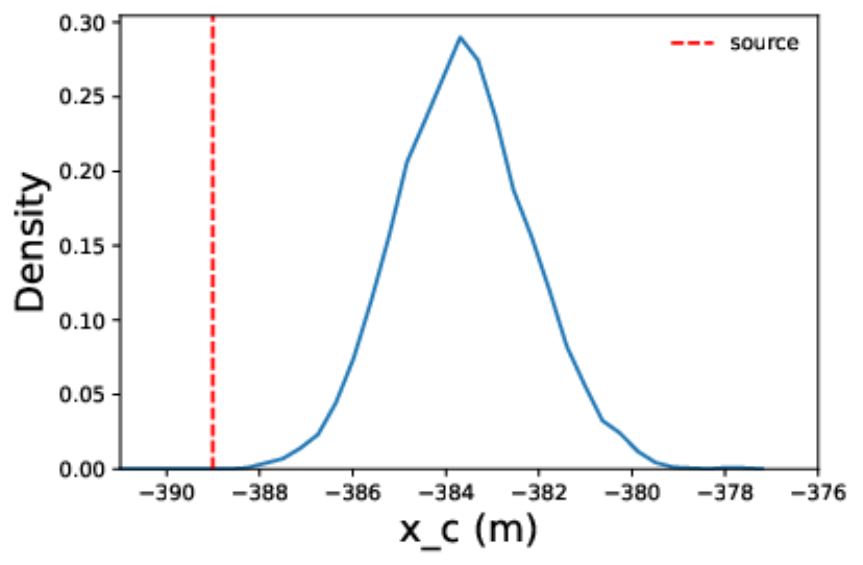}
    \includegraphics[width=6cm]{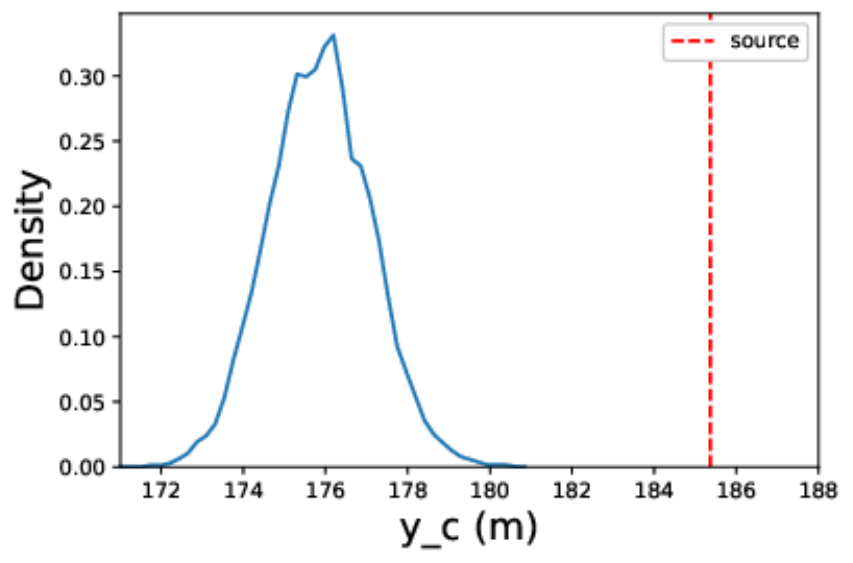}\\
    \includegraphics[width=6cm]{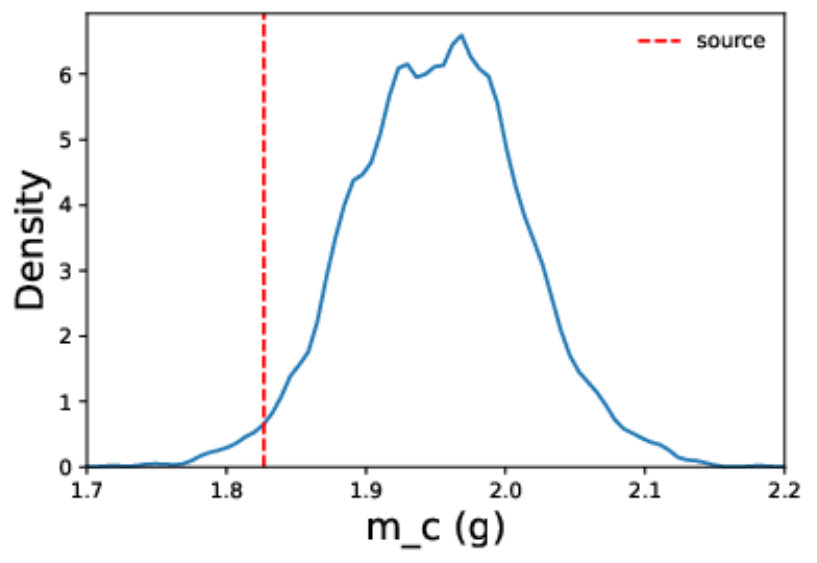}
    \end{tabular}
\caption{Marginal posterior densities for the source location and mass released when accounting for the epistemic uncertainty. The true value of the source location parameters fall outside of the marginal densities, while the true value of the mass released parameter lies just within the marginal density.}
 \label{fig:epistemic}
 \end{figure}

We next examine the robustness of the Bayesian neural network using the same example (the release of 1.83 g of Cs-137 from $(-389.00 \hspace{1mm} \text{m}, 185.37 \hspace{1mm} \text{m})$) from the test dataset. To quantify the epistemic uncertainty of the Bayesian neural network, we predict the source location and mass of the release 100{,}000 times using the same input. The resulting densities for the source location and mass released are plotted in Figure \ref{fig:epistemic}.
 
 Here we observe that the densities are relatively narrow with the true source location falling outside of the densities for $x_c$ and $y_c$ and with the true mass just contained within the density for $m_c$. We note that this quantification of the epistemic uncertainty of the Bayesian neural network does not incorporate the aleatoric uncertainty of our data, which arises from the randomness of the Poisson distribution used to model the randomness inherent to radiation detection.

   \begin{figure}[!ht]
 \centering
 \begin{tabular}{c c}
    \includegraphics[width=6cm]{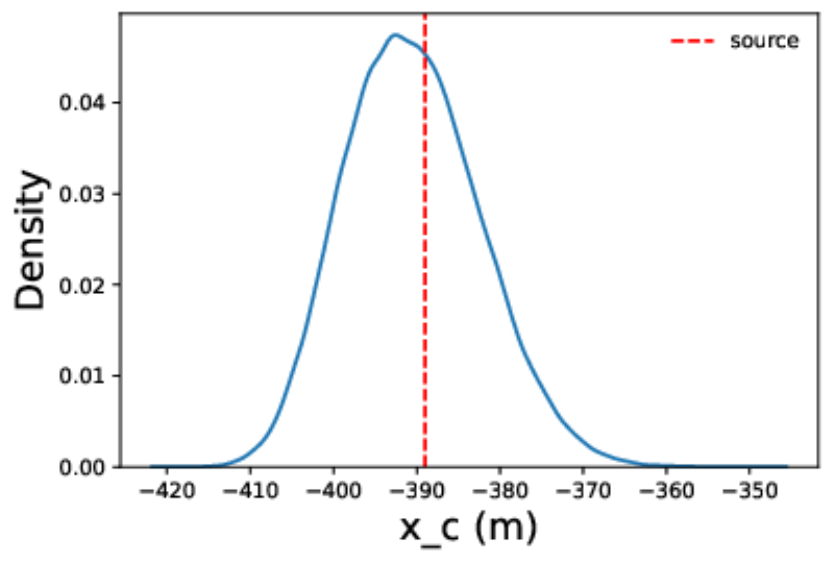}
    \includegraphics[width=6cm]{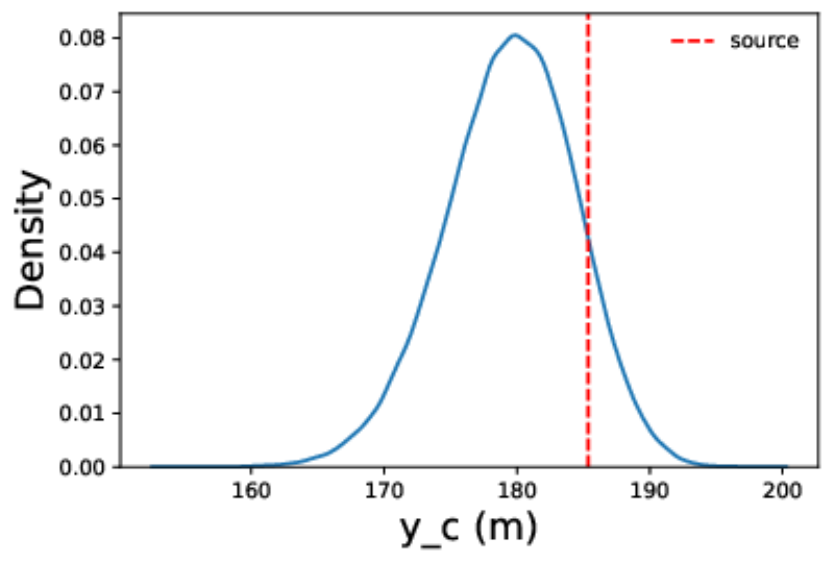}\\
    \includegraphics[width=6cm]{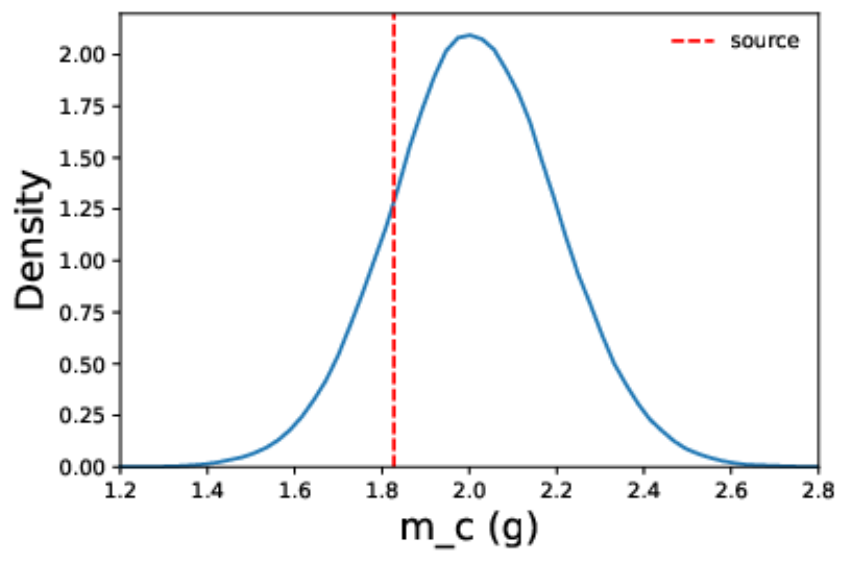}
    \end{tabular}
\caption{Marginal posterior densities for the source location and mass released when accounting for the epistemic and aleatoric uncertainty. The true value of the source location and mass released parameters $(-389.00 \hspace{1mm} \text{m}, 185.37 \hspace{1mm} \text{m}, 1.83 \hspace{1mm} \text{g})$ lies within the marginal densities.}
\label{fig:aleatoric}
 \end{figure}
 
 We account for this randomness, or aleatoric uncertainty of the data, by sampling more detector measurements using the same true release parameters (mass released, source location, wind speeds), and subsequently using the trained Bayesian neural network to predict the source location and mass released for the new sets of detector measurements. To compare with the epistemic uncertainties, we sampled 100{,}000 measurements for each detector and then perform inference using the Bayesian neural network. The resulting densities for the source location and released mass are plotted in Figure \ref{fig:aleatoric}.
 
 When we account for the aleatoric (random) uncertainty of the detector measurements, in addition to the epistemic uncertainty, we observe that the densities for each of the predicted parameters become much wider, and each contains the true value of the release parameters. Comparing these marginal densities to those produced by DRAM as plotted in Figure \ref{fig:densities}, we note that each is slightly more narrow than the corresponding density constructed by DRAM. We note that both DRAM and Bayesian neural network densities overestimate the mass released $m_c$.

 \begin{figure}[!ht]
 \centering
 \begin{tabular}{c c}
    \includegraphics[width=6cm]{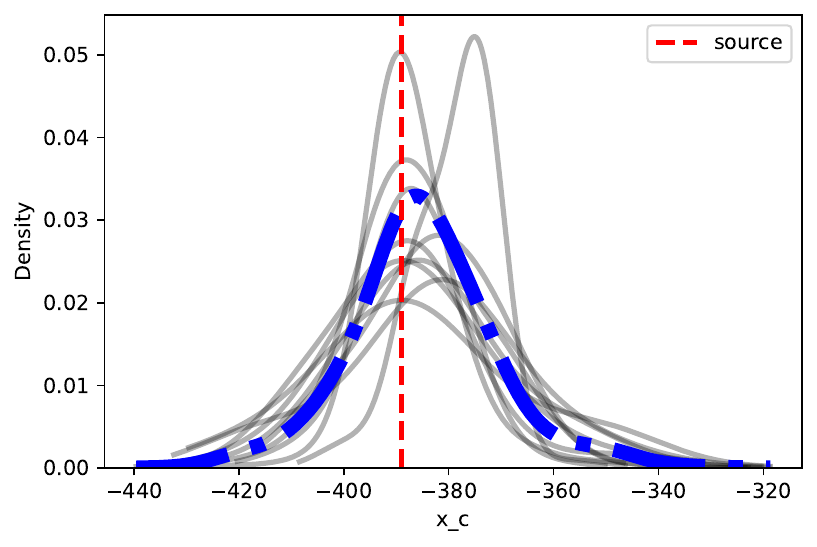}
    \includegraphics[width=6cm]{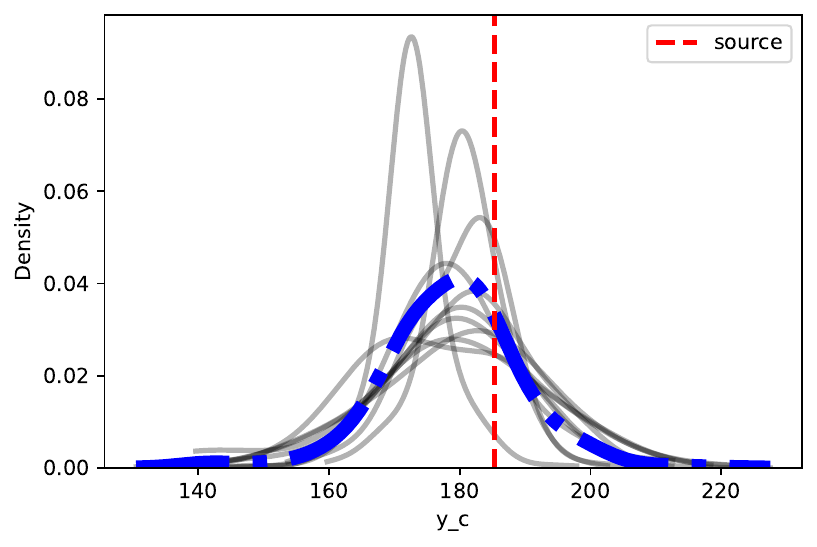}\\   \includegraphics[width=6cm]{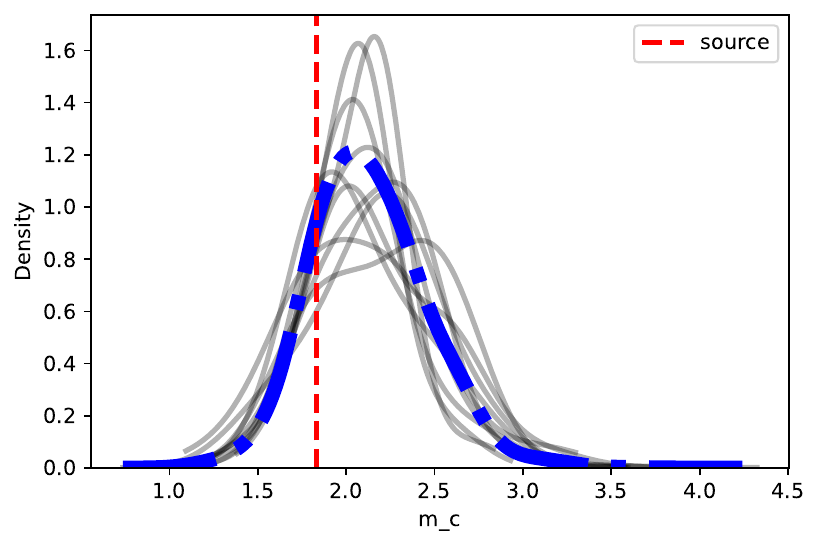}
    \end{tabular}
\caption{Marginal posterior distributions produced by 10 additional DRAM runs. A Gaussian KDE which accounts for all of the data is plotted to illustrate the general trend of the posterior densities.}
\label{fig:DRAM_statistics}
 \end{figure}

 \begin{figure}[!ht]
 \centering
 \begin{tabular}{c c}
    \includegraphics[width=6cm]{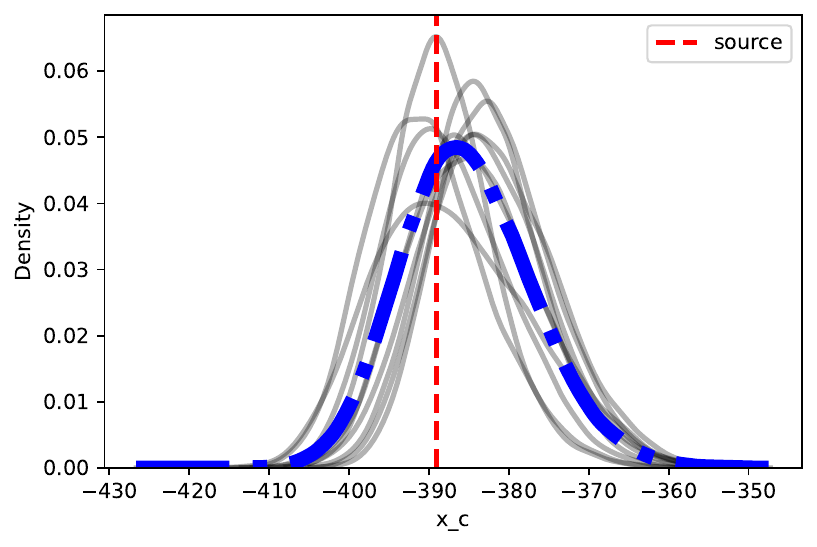}
    \includegraphics[width=6cm]{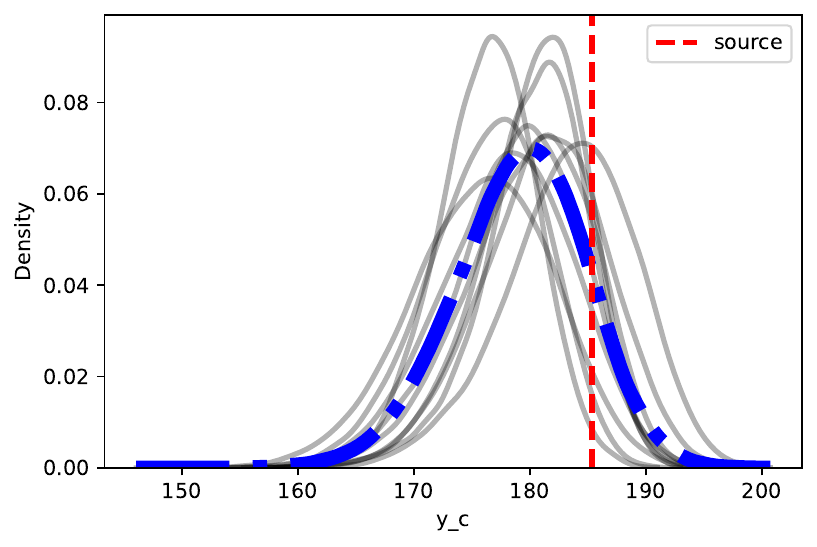}\\
    \includegraphics[width=6cm]{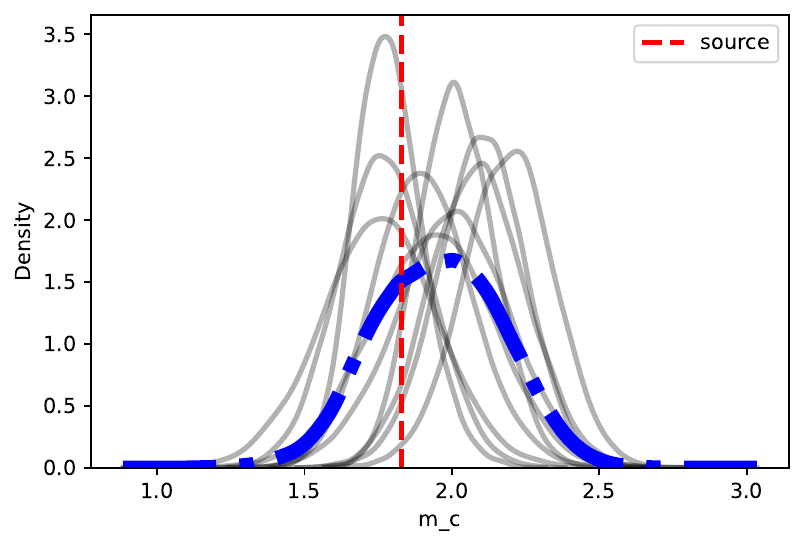}
    \end{tabular}
\caption{Marginal posterior distributions produced by 10 separately trained BNNs. A Gaussian KDE which accounts for all of the data is plotted to illustrate the general trend of the posterior densities.}
\label{fig:BNN_statistics}
 \end{figure}

\begin{figure}[!ht]
 \centering
 \begin{tabular}{c c}
    \includegraphics[width=6cm]{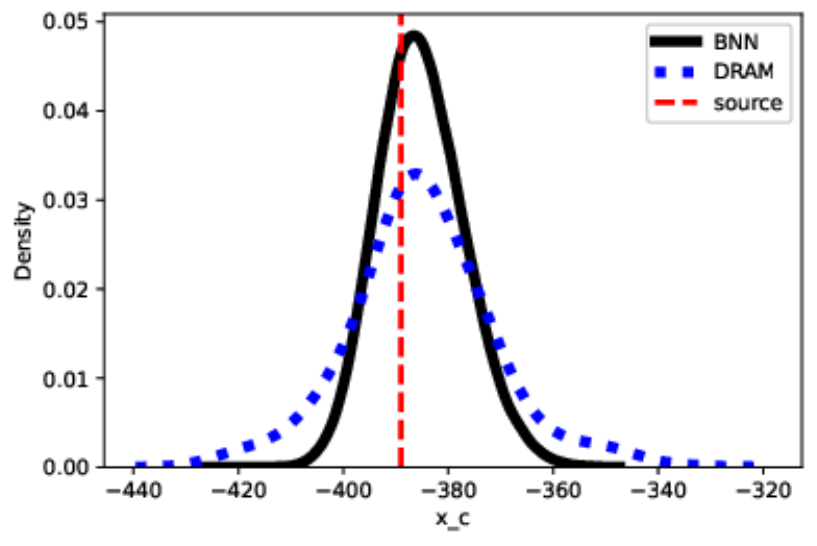}
    \includegraphics[width=6cm]{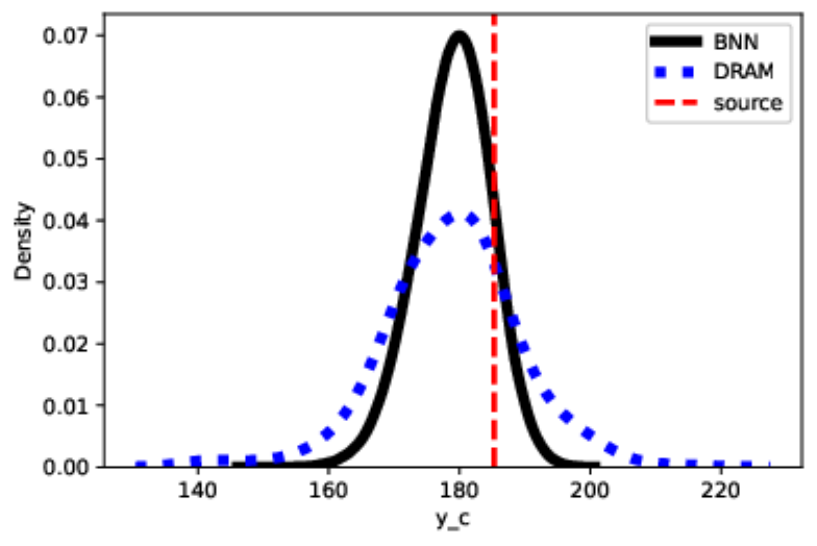}\\
    \includegraphics[width=6cm]{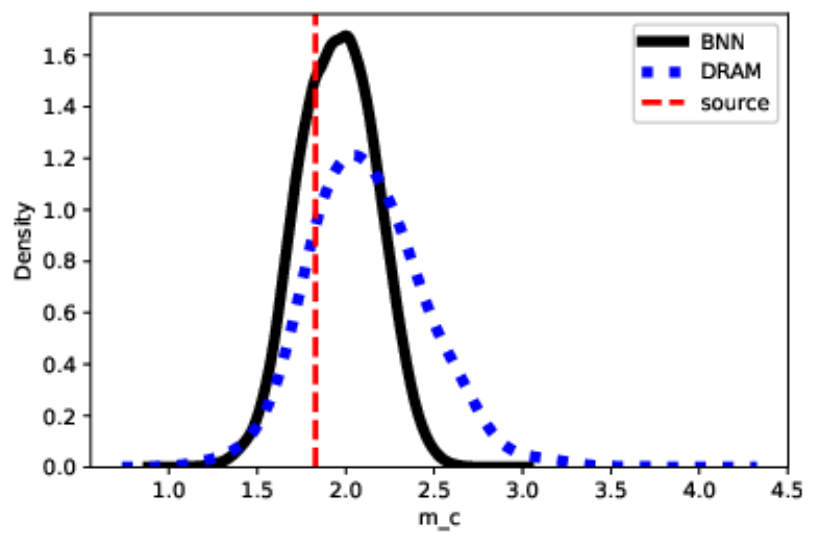}
    \end{tabular}
\caption{Comparison of the KDEs for the distributions produced by DRAM and the Bayesian neural networks. The true parameter values $(-389.00 \hspace{1mm} \text{m}, 185.37 \hspace{1mm} \text{m}, 1.83 \hspace{1mm} \text{g})$ fall within the KDEs for the distributions produced by DRAM and the Bayesian neural networks.}
\label{fig:BNN_DRAM_comparison}
 \end{figure}

  To verify these trends, we conducted additional DRAM and Bayesian neural network testing. We performed 10 additional runs of DRAM for the considered test problem. For the Bayesian neural network, 10 separate networks were trained using the training data. Each network was then utilized to construct densities for the test problem. The resulting densities for DRAM are plotted in Figure \ref{fig:DRAM_statistics}. The resulting densities for the Bayesian neural networks are plotted in Figure \ref{fig:BNN_statistics}. Both figures include a Gaussian kernel density estimate that accounts for points in each of the 10 densities to provide an idea of what this distribution will look like on average. These kernel density estimations (KDEs) are compared directly in Figure \ref{fig:BNN_DRAM_comparison}. 
  
  From these figures we confirm the previous trends that we noted. The DRAM posterior densities are wider than the Bayesian neural network densities, which may signify that the Bayesian neural network is slightly overconfident in its uncertainty predictions. We also observe that whereas both methods generally overestimate the amount of mass released, the Bayesian neural network does so far less. We also note in Figure \ref{fig:BNN_statistics} the significantly different posterior densities constructed for the mass released by the Bayesian neural network. This, along with the limitation of DRAM to identify the amount of mass released, empirically confirms that the mass parameter is more challenging to infer than the source location, which is expected.
  
 We also note the similarity in the width of the densities for $x_c$ and $y_c$ when compared to the histograms of estimated probabilities obtained from the categorical classification neural network. We observe that these histograms are only slightly wider than the densities obtained using the Bayesian neural network when considering both the aleatoric and epistemic uncertainties. Hence the more naive approach of the categorical classification neural network performs well for this problem. Finally, we note the computational time required to compute a 10{,}000 prediction density, for comparison to DRAM's computational time. A 10{,}000 prediction density accounting for the epistemic uncertainty was computed in 10.7 s. When accounting for both the aleatoric and epistemic uncertainties the densities required 10.9 s to compute, which is significantly faster than the approximately 1600 seconds required to run DRAM for this problem. We note that this speedup is due to the construction of the density relying on the computational cost of the neural network model rather than the computational expense of sampling the transport and radiation detection models to construct a posterior parameter distribution.
 
 \section{Conclusion and Future Work}
 \label{sec:conclusion3}
 We developed fast neural networks, which were capable of accurately inferring the release parameters given the mean wind speeds and detector measurements at a single time. We demonstrated that MCMC techniques, such as DRAM, can be used to infer posterior densities for the parameters and quantify uncertainty; however, they do not provide the capability for real-time implementation. We showed that the regression-based neural network is accurate and able to solve the problem in real-time, but it provides no quantification of the uncertainty. As noted in Section \ref{sec:chap3_dram_results}, this does not provide a mechanism to quantify the accuracy or reliability of the deterministic predictions. To address this, we constructed a categorical classification neural network by binning the data. The categorical classification neural network estimated the probability that the true release location fell within given bins and provided a comparison to the marginal and posterior densities obtained from DRAM. Finally, we constructed a Bayesian regression-based neural network, which predicted the source location and released mass to quantify the epistemic uncertainty of the neural network. The densities for the release parameters, when considering the epistemic uncertainty, proved to be narrow and did not contain the true parameter for 2 of the 3 release parameters. However, when we additionally considered the aleatoric uncertainty of the detector measurements, we obtained wider marginal densities that all contained the true values of the release parameters. These densities were similar to the DRAM predictive posterior distributions, with the Bayesian neural network densities being slightly more narrow. Both exhibited similar properties, however, such as overestimating the mass released $m_c$ parameter. 

There are many directions for future research. First, one can extend this method to 3-D which would require an array of sensors at a range of heights. Another potential research direction is to develop a machine learning technique that can be combined with radiation mapping to quickly localize hotspots of deposited radiation. A useful extension of this method would be to construct a similar neural network capable of incorporating data across multiple times. A further study with a more advanced transport model and domain, such as an urban environment, is another area of future research. In these cases, a larger neural network architecture may be required to resolve the introduced complexities of the transport model and domain.

\pagebreak
\section*{Acknowledgments}

\noindent This research was supported in part by the Air Force Office of Scientific Research (AFOSR) through the grant AFOSR FA9550-18-1-0457.  The research of RCS was also supported in part by the National Science Foundation through the grant DMS-2053812.

\section*{Disclosure Statement}
\noindent The authors report there are no competing interests to declare.

\bibliography{Edwards_Paper}

\end{document}